\documentclass[runningheads]{llncs}
\usepackage{graphicx}
\usepackage[dvipsnames]{xcolor}
\usepackage[numbers,sort&compress]{natbib}
\usepackage[dvipsnames]{xcolor}
\usepackage{multirow}
\usepackage{epstopdf}
\usepackage{floatrow}
\newfloatcommand{capbtabbox}{table}[\FBwidth]
\usepackage{tabularx}
\usepackage{graphicx}
\usepackage{soul}
\usepackage{adjustbox}
\usepackage[hyphens]{url}
\usepackage{hyperref}
\usepackage{vwcol}
\usepackage[inline]{enumitem}
\setlist[enumerate,1]{label=\textbf{(\arabic*)}}
\let\llncssubparagraph\subparagraph
\let\subparagraph\paragraph

\usepackage[compact]{titlesec}
\titlespacing{\section}{0pt}{2ex}{1ex}
\titlespacing{\subsection}{0pt}{1ex}{0.2ex}
\titlespacing{\subsubsection}{0pt}{0.1ex}{1ex}

\let\subparagraph\llncssubparagraph

\begin{document}
\title{Atlas: A Dataset and Benchmark for E-commerce Clothing Product Categorization}

%
%
\author{Venkatesh Umaashankar \inst{1}\orcidID{0000-0001-5230-1209} \and
Girish Shanmugam S\inst{2}\orcidID{0000-0003-2805-7503} \and
Aditi Prakash\inst{3}\orcidID{0000-0002-4839-9132}}
\authorrunning{V. Umaashankar et al.}
%
\institute{Ericsson Research, Chennai, India \\ \email{venkatesh.u@ericsson.com}\\ \and
Ericsson Research, Chennai, India \\ 
\email{s.girishshanmugam@gmail.com}\\
\and
University of Colorado, Boulder, Colorado, USA\\
\email{adpr5166@colorado.edu}}
\titlerunning{Atlas: A Dataset for E-commerce Clothing Product Categorization}
\maketitle              
\begin{abstract}
In E-commerce, it is a common practice to organize the product catalog using product taxonomy. This enables the buyer to easily locate the item they are looking for and also to explore various items available under a category. Product taxonomy is a tree structure with 3 or more levels of depth and several leaf nodes. Product categorization is a large scale classification task that assigns a category path to a particular product. Research in this area is restricted by the unavailability of good real-world datasets and the variations in taxonomy due to the absence of a standard across the different e-commerce stores. In this paper, we introduce a high-quality product taxonomy dataset focusing on clothing products which contain 186,150 images under clothing category with 3 levels and 52 leaf nodes in the taxonomy. We explain the methodology used to collect and label this dataset. Further, we establish the benchmark by comparing image classification and Attention based Sequence models for predicting the category path. Our benchmark model reaches a micro f-score of 0.92 on the test set. The dataset, code and pre-trained models are publicly available at \url{https://github.com/vumaasha/atlas}. We invite the community to improve upon these baselines.

\end{abstract}
\section{Introduction}

With the Internet revolution, E-commerce has become a major platform for selling products to customers. E-commerce stores host a collection of products ranging from electronics to fashion apparel to grocery. A well-organized E-commerce store lets customers navigate through the website with ease and locate the product they are looking for. Unlike a traditional retail store where you can walk in and seek assistance, online retailers rely on their product catalog or categorization to assist shoppers to find their desired product. Product taxonomy is a tree structure with multiple top and intermediate levels, ending in leaf nodes. Taxonomy classification is the process of assigning a category path to a particular product in the taxonomy tree. E-commerce sites use hierarchical taxonomies to organize products
from generic to specific classes where each level provides more specific details about the product than the previous level. For example, \textbf{\textit{Clothing \& Accessories \textgreater Men \textgreater Winterwear \textgreater Sweatshirts \& Hoodies}}. These classification levels are important for an E-commerce store to perform operations such as search, catalog building, recommendation, which thereby hugely influence customer satisfaction and revenue of e-commerce sites. Currently, most of these product classification mechanisms rely on sellers to provide correct details. Each E-commerce store has its own product taxonomy and a seller typically sells in multiple stores. This implies that the seller has to perform such categorizations manually multiple times.
Automating this has potential benefits of reduced costs and better catalog quality. Our key contributions in this paper are:
\begin{enumerate*}
    \item Developed a clean and rich clothing product taxonomy dataset containing 186,150 images and their corresponding product titles which maps to 52 category paths.
    \item Proposed a methodology to collect large scale product taxonomy dataset which can be easily extended to categories other than Clothing.
    \item Trained and compared two benchmark models (Image classification and Attention based Seq to Seq model) that predicts the category path from the product image. Our best model reached an f-score of 0.92 on the test set.
    \item The dataset, source code and pre-trained models are made publicly available\footnote{\url{https://github.com/vumaasha/Atlas}} to encourage future research in this area.
\end{enumerate*}

 The rest of the paper is organized as follows. Related literature is reviewed in Section \ref{sec:relatedwork}. In Section \ref{sec:dataset} we explain the methodology that we used to develop the \textbf{\textit{Atlas}} dataset. In Section \ref{sec:approach} we build our benchmark models and explain our model architecture. In Section \ref{sec:training} we provide our training setup and meta parameters to facilitate reproducible research. In Section \ref{sec:results} we summarize our results. Finally, in Section \ref{sec:conclusion} we conclude and provide details about possible directions for future work.

\section{Related Work}
\label{sec:relatedwork}

A clean and detailed product taxonomy offers several benefits to both the E-commerce store and its customers. However, creating, maintaining or adapting an existing categorization standard is not an easy task.  Still, most of E-commerce stores want to have flexibility in the way they organize their catalog and create their product taxonomy.

Initially, techniques from
information retrieval and machine learning were applied to solve the problem of product categorization. GoldenBullet \cite{ding2002goldenbullet} is a software environment targeted to automatically classify the products, based on their original descriptions and existent classification standards (such as UNSPSC). It integrates different classification algorithms like Vector space model (VSM), K-nearest neighbor and Naive-Bayes classifier algorithms and some natural language processing techniques to pre-process data. \cite{dumais2000hierarchical} approached product categorization as a hierarchical text classification task. They proposed two different approaches of building separate classifiers for each level in the hierarchy and a flat classifier that directly predicts the leaf level assignment of a document. They used Support Vector Machine (SVM)  classifiers for evaluating both the approaches. \cite{kozareva2015everyone} presented a simple linear classifier based approach for product categorization using mutual information and LDA based features. In general, the computational complexity involved in some of these traditional machine learning techniques is well beyond linear with respect to the number of training examples, features, or classes. The
scale of the E-commerce product categorization requires algorithms capable of processing a huge volume of training data in a reasonable time, capable of handling a large
number of classes and also capable of making fast real-time predictions.\citep{shen2012large}.

The remarkable progress made in the field of deep learning in recent years has provided a better way to approach this problem. \cite{das2017web} has done a detailed study of using Convolutional Neural Networks (CNN) for the product categorization task. They used the Amazon product dataset provided by \cite{mcauley2015image} and text features such as product titles, navigational breadcrumbs, and list price. \cite{ha2016large} used multiple Deep Recurrent Neural Network (RNNs) and generated features from the text metadata. In recent times, Sequential model-based approaches have been widely used for product categorization. \cite{DBLP:conf/sigir/HiramatsuW18} modeled product categorization as a Sequence to Sequence learning, they used product titles which are a sequence of words as input and predicted the category path as a sequence of category levels in the product taxonomy. 
        
Due to the availability of large high-quality image datasets, the field of Image classification \citep{krizhevsky2012imagenet} has matured a lot in recent times. Noise and ambiguity is a common problem in textual product titles and description. However, most of the E-commerce products tend to have decent product images, this leads to a natural choice of using images for product categorization. \cite{bossard2012apparel}, \cite{chen2012describing}, \cite{kiapour2014hipster} and \cite{liu2016deepfashion} applied computer vision techniques for  fashion apparel categorization based on the product images. The closest to our work is by  \cite{li2018unconstrained} where they use Seq to Seq model with product titles as input to predict category paths using an LSTM Decoder and beam search for inference. We extend their work in this paper by using product images instead of product titles as input for product categorization. Similar to \cite{li2018unconstrained}, we learn an Attention based Seq to Seq model. 

\section{Atlas Dataset}
\label{sec:dataset}

Rakuten made a product classification dataset publicly available in Rakuten Data Challenge \cite{lin2018overview}, However, this dataset contains only the product titles and the levels in the taxonomy are represented using numerical IDs instead of plain text. Real world product taxonomy datasets are
not publicly available.  Also, there is no widely adopted industry standard for defining product taxonomies. In addition to these, factors like data size, category skewness, and noisy metadata are limiting further research and practical implementation of large scale product categorization. This motivated us to develop a real-world dataset for product categorization.

We developed a new product categorization dataset called \textbf{\textit{Atlas}}. An E-commerce store typically sells products under several top-level categories such as Electronics, Home \& Kitchen, Clothing, etc. In this paper, we focused only on clothing products, \textcolor{blue}. Our dataset contains data corresponding to 52 products and their title, price, image and category path.

\subsection{Taxonomy Generation}
\label{ssec:taxonomy}


In the E-commerce world, each store has its own taxonomy. For example the category path for 'Jackets' in Flipkart\footnote{\url{https://www.flipkart.com/}} is \textbf{\textit{Clothing \textgreater Men's Clothing \textgreater Winter \& SeasonalWear \textgreater Jackets}} and in Amazon\footnote{\url{https://www.amazon.in/}} it is  \textbf{\textit{Clothing \& Accessories \textgreater Men \textgreater Jackets}}. Treating them as different category paths will lead to noisy taxonomy and training data. We designed our taxonomy based on the similarities in the taxonomy structures across the different e-commerce retailer websites. Our taxonomy is organized to a maximum depth of 3 levels which can assist the consumer to reach their product in not more than 3 clicks. The process of building our taxonomy involved three steps. First, we analyzed and listed the taxonomy structures of popular products, niche, and premium clothing products across different e-commerce stores. 
Next, we identified the common category paths up to the third level across these websites. Finally, clothing that had the same category paths until the third level were clubbed together irrespective of the dissimilarity in the deeper levels. For example, \textbf{\textit{Women \textgreater Ethnic Wear \textgreater Salwar Kameez \textgreater Bollywood}} and \textbf{\textit{Women \textgreater Ethnic Wear \textgreater Salwar Kameez \textgreater Anarkali}} are grouped together under the category \textbf{\textit{Women \textgreater Ethnic Wear \textgreater Salwar Kameez}} as they have similar category paths until the third level beyond which it branches into different nodes. Our final taxonomy tree is not an exhaustive list of all clothing categories but cover popular Western Wear and niche Ethnic Wear especially from the Indian Clothing collection. Each of the categories in the taxonomy tree have a maximum depth of 3 levels and totals to 52 category paths. Our taxonomy tree and a  few sample products from some of the categories in our dataset can be found here \footnote{\url{https://github.com/vumaasha/Atlas/tree/master/dataset\#11-taxonomy-generation}}. 

\subsection{Data Collection}
We crawled the product listings from popular Indian E-commerce stores. We manually created a mapping of the store's category path that maps to a category path in our taxonomy. We used web scraping tools Scrapy and Selenium. The crawlers extract the information from the HTML content of the product page using CSS selectors. We extracted the  \textit{product title, breadcrumb, image and price} corresponding to each product in the product listings. The attributes extracted are stored in JSON format. 


\begin{figure}
    \centering
    \includegraphics[width=\textwidth]{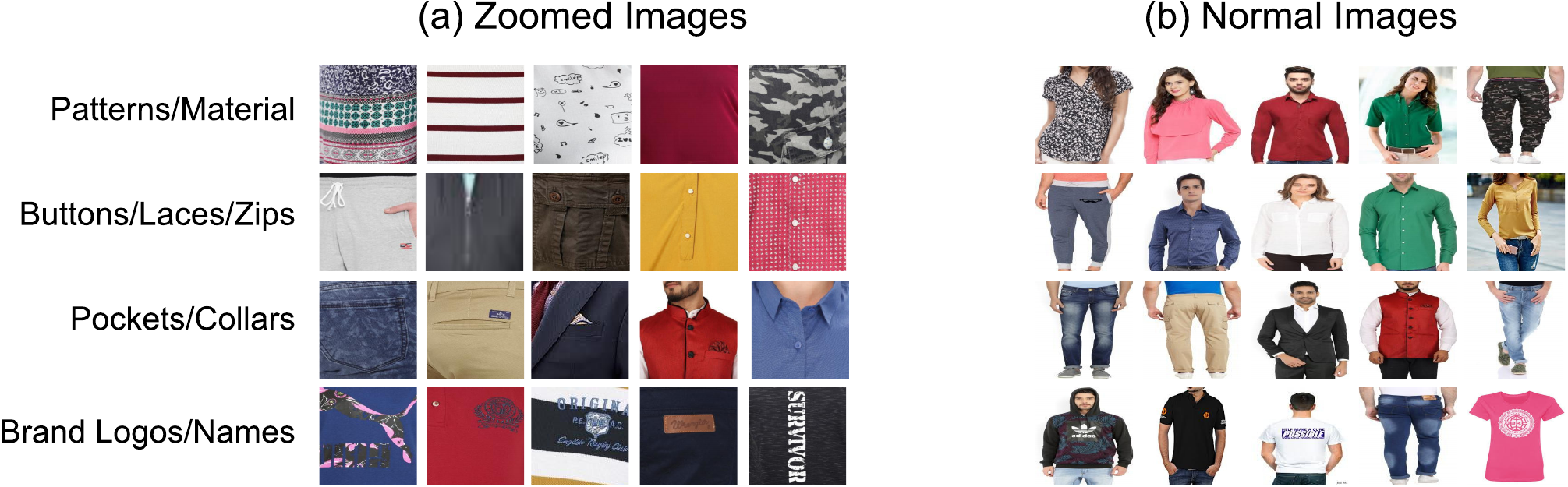}
    \caption{Examples of (a) Zoomed(dirty) and (b) Normal(clean) images from our \textbf{\textit{Atlas}} dataset. The Zoomed images show close-ups of the apparel or cropped versions of the image that make it difficult to recognize the product, whereas the Normal images show figures with the entire product visible.}
    \label{fig:normal_zoomed}
\end{figure}

\begin{figure}
    \centering
    \includegraphics[width=\textwidth]{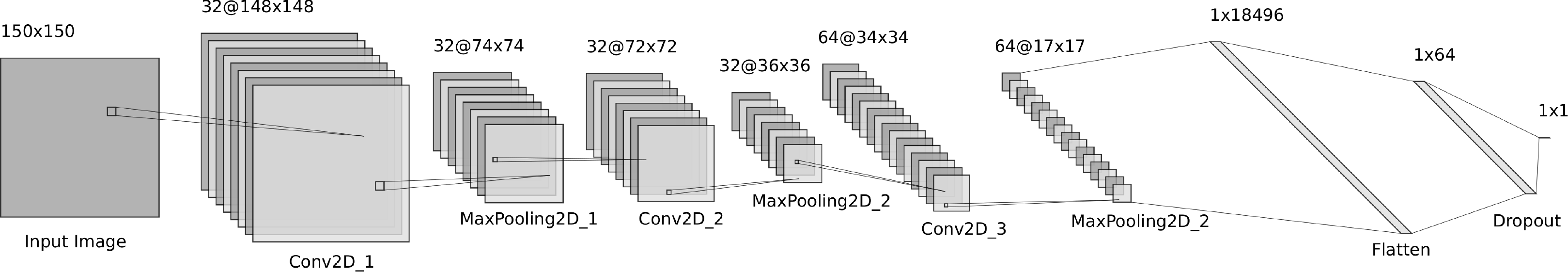}
    \caption{Architecture of Zoomed Vs Normal Model}
    \label{fig:zoomed_vs_normal_cnn}
\end{figure}

\subsection{Data Cleaning}
It is typical for an E-commerce store to show several images for a single product. Out of these images, not all the images are necessarily a good representative image of the product. Some images might display packaging, installation instructions, etc. In the case of clothing, we found that many product listings also included zoomed in images that display intrinsic details such as the texture of the fabric, brand labels, button, and pocket styles. Without the context of the product listing, it would be even hard for a human to identify the corresponding product. Including these zoomed in images would drastically affect the quality of the dataset. To find and remove these noisy images manually would take considerable time and effort. We modeled this as a binary classification task (Zoomed Vs Normal Images) and compared Linear SVM with simple 3 layer CNN (Figure. \ref{fig:zoomed_vs_normal_cnn}) based classification models. We prepared the training data by visual inspection. We segregated noisy and high-quality images into two different folders by looking at the thumbnails of hundreds of product images in a go. Our models were trained on 6005 normal images and 1054 zoomed images and the performance metrics on the test are shown in the Table. \ref{tab:zvsn_metrics}. We used computer vision based features such as contors and histogram of gradients as input for our LinearSVM. We automated the process of filtering out the noisy images using the CNN model due to its superior performance compared to that of LinearSVM.

\begin{table}
    \centering
  \begin{tabular}{|p{2cm}|c|c|c|c|c|c|}
    \hline
    \multirow{2}{*} & \multicolumn{3}{c|}{\textbf{CNN}} & \multicolumn{3}{c|}{\textbf{SVM}}\\
    \cline{2-7}
    & \textbf{precision} & \textbf{recall} & \textbf{f-score} & \textbf{precision} & \textbf{recall} & \textbf{f-score}\\
    \hline
    Normal & 0.99 & 0.99 & 0.99 & 0.91 & 0.99 & 0.95 \\ \hline
    Zoomed & 0.95 & 0.95 & 0.95 & 0.86 &
    0.48 & 0.62 \\ \hline
    \textbf{Average} & 0.98 & 0.98 & 0.98 & 0.91 & 0.91 & 0.90 \\ \hline
    \end{tabular}
    \caption{Metrics for the models used to predict Zoomed Vs Normal images}
    \label{tab:zvsn_metrics}
\end{table}


\section{Benchmark models for Product Categorization}
\label{sec:approach}
\subsection{Resnet34 based Image Classification}

We use the cnn\_learner available in fast.ai implementation to train our Image classification model. This uses Resnet34 architecture as the backbone of the model, which is followed by AdaptiveConcatPool2d, Flatten and 2 blocks of [nn.BatchNorm1d, nn.Dropout, nn.Linear, nn.ReLU] layers. The first block will have a number of inputs inferred from the backbone arch Resnet34 (512) and the second one will have a number of outputs equal to the number of classes (52) without nn.ReLU activation.

\subsection{Attention based Seq to Seq Model}

\begin{figure}
    \centering
    \includegraphics[width=0.8\linewidth]{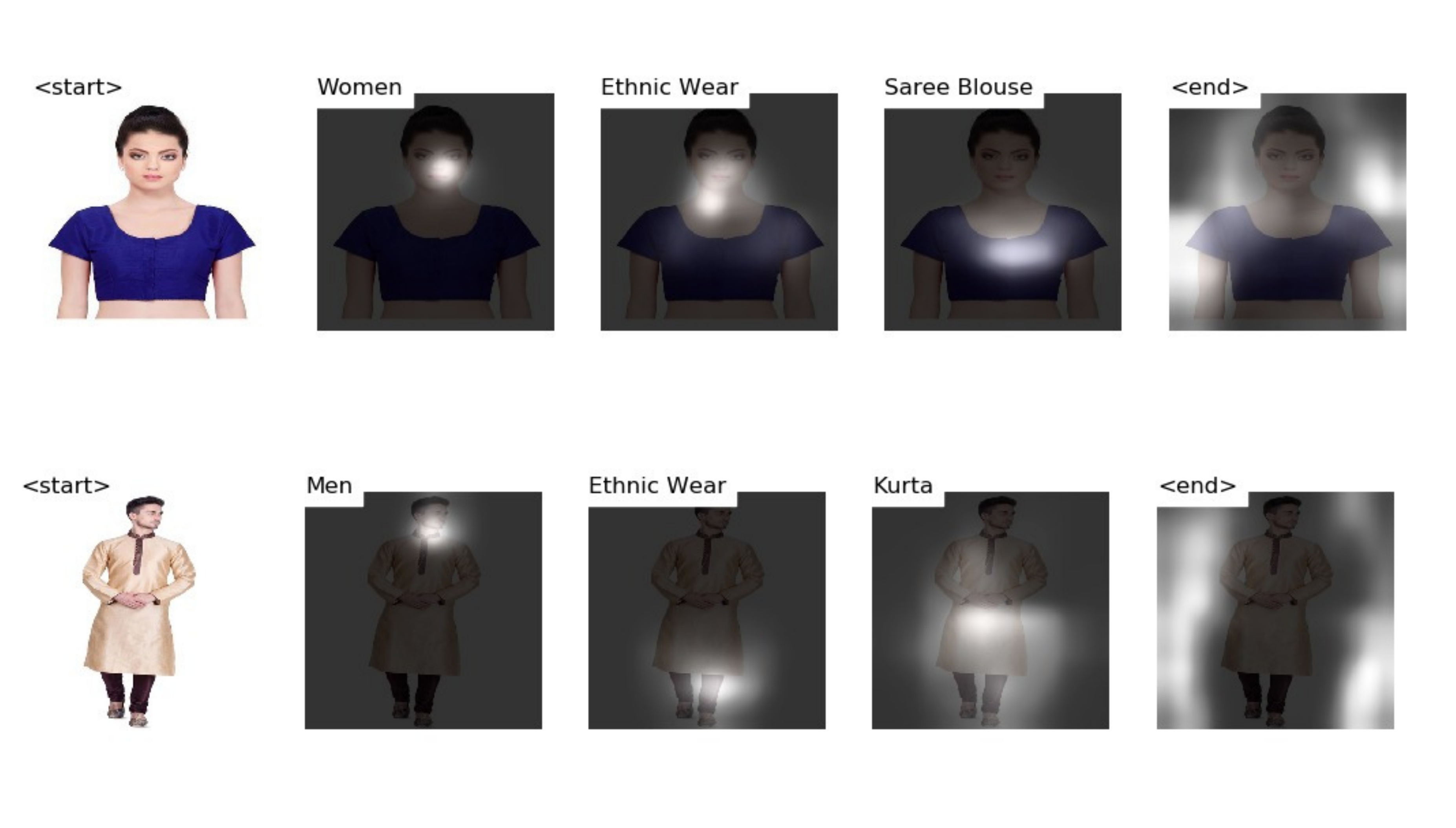}
    \caption{A sample of category paths predicted on test dataset by our model. We can observe how the Attention focuses on different sections of the image while generating each category level. For example, the face is being focused to predict the first category level - gender.}
    \label{fig:sample_predictions}
\end{figure}

We approach the product categorization problem as a sequence prediction problem by leveraging the dependency between each level in the category path. We use Attention based Encoder-Decoder Neural Network architecture to generate sequences. The Encoder is a 101 layered Residual Network(ResNet) trained on the ImageNet classification task which converts the input image to a fixed size vector. The Decoder is a combination of Long Short-Term Memory(LSTM) along with Attention Network which combines the Encoder output and Attention weights to predict category paths as sequences. Our architecture is similar to works by \cite{wu2016google} and \cite{xu2015show} used for Neural Machine Translation and image captioning respectively. We extended the source code from pytorch tutorial to image captioning repository by Sagar Vinodababu\footnote{\url{https://github.com/sgrvinod/a-PyTorch-Tutorial-to-Image-Captioning}}. Figure \ref{fig:sample_predictions} shows some of the category paths generated by our model on test images which are not seen during training or validation. From this figure, it can be clearly seen that to generate each category level our model focuses on different parts of the image. To predict the first category level, which is the gender, our model has focused on the face and it has focused on the actual region of the clothing products to predict the next category levels, 'SareeBlouse' and 'Kurta'.

\subsubsection{Encoder and Decoder}

\begin{figure*}
\begin{multicols}{2}
    \includegraphics[width=\linewidth]{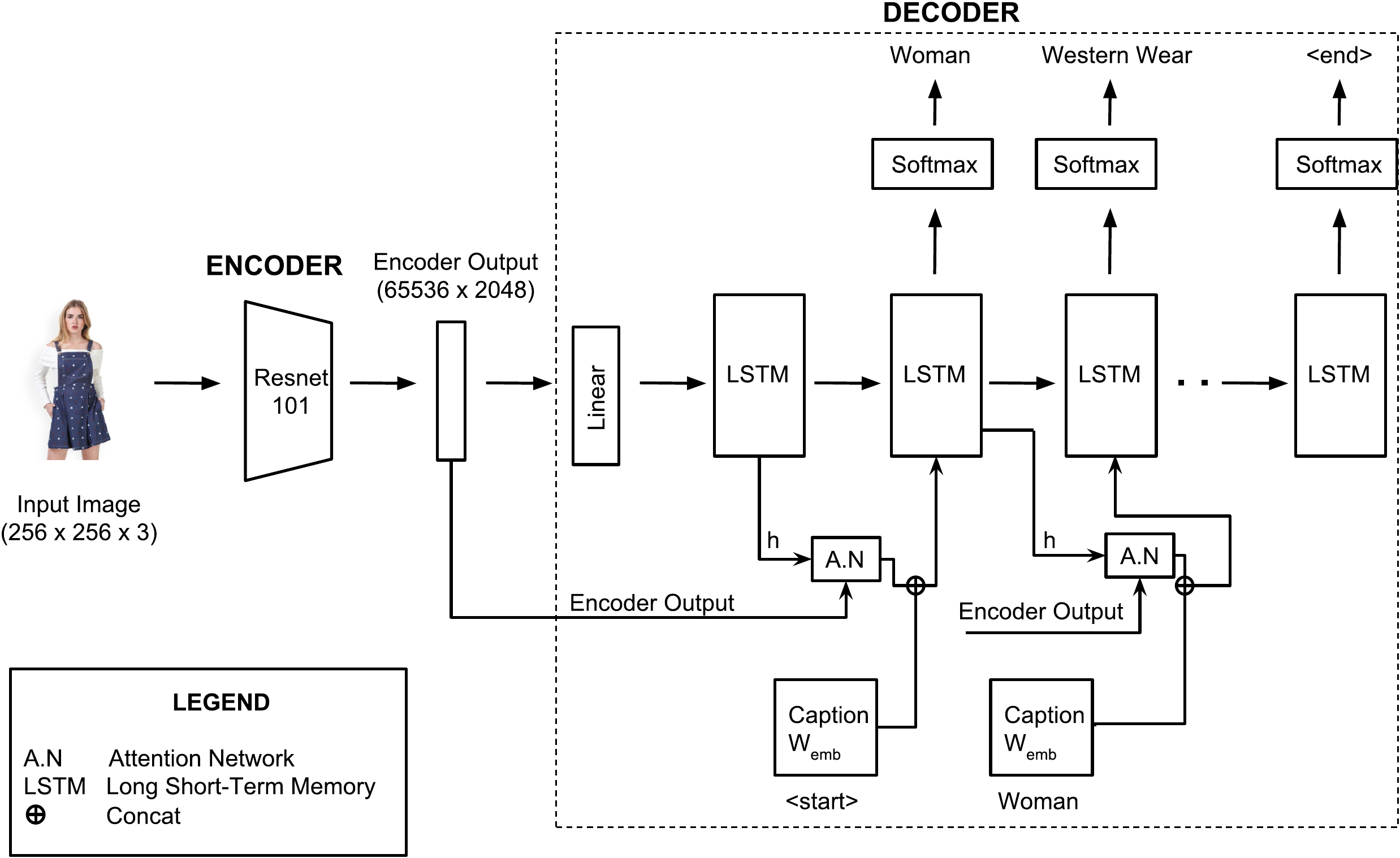}
    \vfill\columnbreak
    \includegraphics[width=\linewidth]{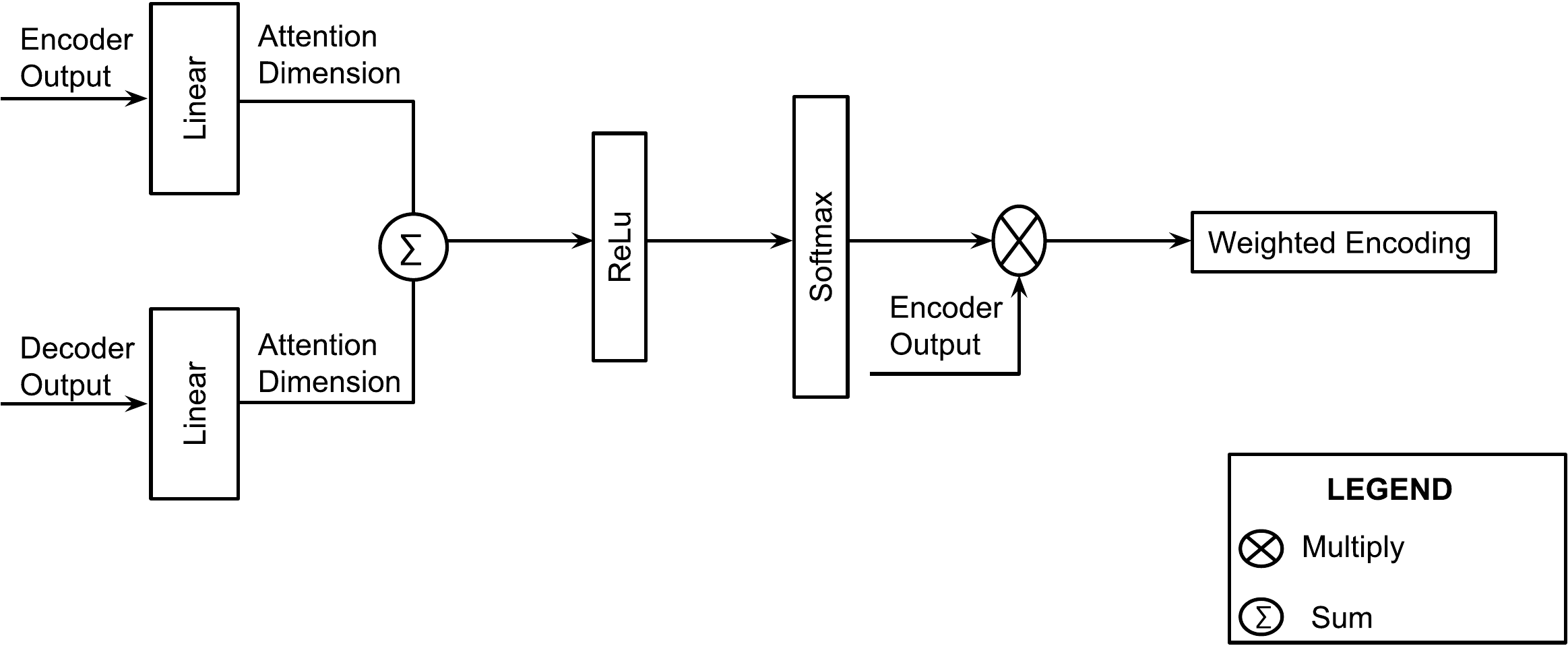} 
\end{multicols}
\caption{Encoder - Decoder with Attention Network}
\label{fig:encoder_decoder}
\end{figure*}

In Encoder, we use Convolutional Neural Network (CNN) to produce fixed size vectors. The images in \textbf{\textit{Atlas}} dataset have different dimensions as they were collected from different sources. All these images are resized to have a uniform dimension of 150*150 pixels before being fed as input to the Encoder. The input images are then represented by the 3 color channels of RGB values. The Encoder uses a 101 layered Residual Network pre-trained on the ImageNet classification task which is shown in Figure \ref{fig:encoder_decoder}. As we use the Encoder only to encode images and not for classifying them, we remove the last two layers (linear and pooling layers) from the ResNet-101 model proposed by \cite{he2016deep}. The images are resized to a fixed size by adding a 2D adaptive average pooling layer which enables the Encoder to accept images of variable sizes. The final encoding produced by the Encoder will have the dimensions: batch\_size,14,14, 2048.

Recurrent Neural Networks(RNN) are popular for sequential classification task as it considers both the current input and the learnings from the previously received inputs for prediction. Usually, RNN's have short term memory but when combined with Long Short-Term Memory (LSTM) Network they have long term memory as LSTM’s contain their information in a memory. We have a stacked LSTM Network along with Attention in our Decoder which is shown in Figure \ref{fig:encoder_decoder}. 

The Attention Network shown in Figure \ref{fig:encoder_decoder}, learns which part of the image has to be focused to predict the next level in the category path while performing the sequence classification task. The Attention Network generates weights by considering the relevance between the encoded image and the previous hidden state or previous output of the Decoder. It consists of linear layers which transform the encoded image and the previous Decoder's output to the same size. These vectors are summed together and passed to another linear layer. This layer calculates the values to be Softmaxed and then passes the values to a ReLU layer. A final softmax layer calculates the weights \textit{alphas} of the pixels which add up to 1. If there are P pixels in our encoded image, then at each time step t,
\begin{equation}
\sum_p^P\alpha_{p,t} = 1
\end{equation}
 We use a weighted average across all the pixels instead of a simple average so that the important pixels are assigned greater weights. 
 
 The Decoder receives the encoded image from the Encoder using which it initializes the hidden and cell state of the LSTM model through two linear layers. Two virtual category levels \textbf{\textless start\textgreater}  and \textbf{\textless end\textgreater} which denote the beginning and end of the sequence are added to the category path. The Decoder LSTM uses teacher forcing proposed by \cite{williams1989learning} for training. The Decoder uses a \textbf{\textless start\textgreater}  marker which is considered to be the zeroth category level. The  \textbf{\textless start\textgreater}  marker along with the encoded image is used to generate the first-top level of the category path. Subsequently, all other levels are predicted using the sequence generated so far along with the Attention weights. An \textbf{\textless end\textgreater}  marker is used to mark the end of a category path. The Decoder stops decoding the sequence further as soon it generates the  \textbf{\textless end\textgreater} marker. At each time step, the Decoder computes the weights and Attention weighted encoding from the Attention Network using its previous hidden state. Another linear layer is added to create a sigmoid-activated gate and the Attention weighted encodings are passed through it and concatenated with the embedding of the previously generated category path and fed into the LSTM Decoder to generate the new hidden state which is also the next predicted level. The next level is predicted using a final softmax layer from the hidden state of the Decoder. The softmax layer transforms the hidden state into scores which are stored for further utilization in beam search for selecting 'k' best levels.

\section{Training}
\label{sec:training}

\subsection{Model hyperparameters}

\subsubsection{Zoomed Vs Normal LinearSVM Model}
We trained LinearSVM available in Scikit-learn with C set to 0.0001, class weight set to 'balanced' using hinge loss. The optimal C value was identified using a grid search.

\subsubsection{Zoomed Vs Normal CNN Model}
We trained for 10 epochs using Binary CrossEntropy as loss function, RMSProp Optimizer with a learning rate set to 0.001, rho set to 0.9 and decay set to 0.0.

\subsubsection{Resnet34 based Image Classification}
We trained for 17 epochs using Categorical CrossEntropy as loss function and Leslie Smith's one cycle policy \cite{smith2017cyclical} for choosing the learning rate. We used early stopping to terminate the training process when the decrease in validation loss is less than 0.001 for 3 consecutive epochs.

\subsubsection{Attention based Seq to Seq Model}
We trained our model in GPU for 3 epochs with a batch size of 128 and dropout rate as 0.5 after which the validation accuracy stop improving. We used Adam optimizers with a learning rate of 1e-4 and 4e-4  for Encoder and Decoder respectively. We picked the beam width as 5 based on our experiments. Regularization parameter for doubly stochastic Attention was set to 1 and gradient clipping was set to an absolute value of 5. The pre-trained model can be downloaded from here\footnote{\url{https://goo.gl/forms/C1824kjmbuVo7H6H3}}.

\subsection{Hardware}

\begin{enumerate*}
\item  Nvidia GPU GEFORCE GTX 1080 Ti 11GB RAM
\item Intel\textsuperscript{\textregistered} Xeon\textsuperscript{\textregistered} Processor E5-2650 v4 30M Cache, 
2.20 GHz, 12 Cores, 24 Threads
\item 250 GB RAM
\item CentOS 7
\end{enumerate*}

\section{Results}
\label{sec:results}
\begin{figure}
    \centering
    \includegraphics[width=1.0\textwidth]{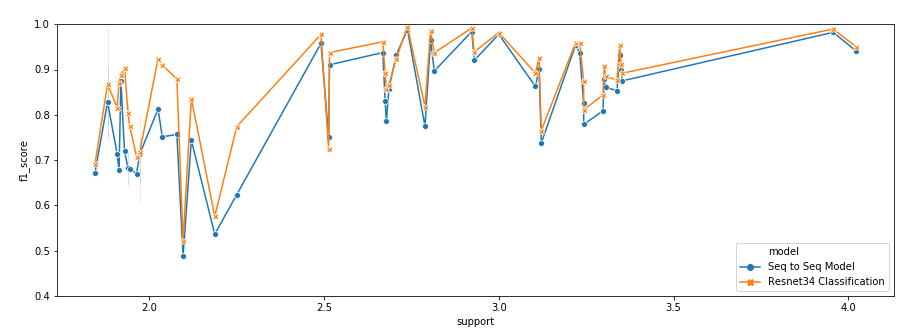}
    \caption{F-scores of our benchmark models over leaf level categories ordered by their sample size. Note that the sample size in \textit{x} axis is in log scale}
    \label{fig:metrics_comparison}
\end{figure}

We evaluated the proposed model on our dataset having 186,150 clothing images and their category paths. We split our dataset into train, validation and test sets similar to the splits used in the work by \cite{karpathy2015deep}. Stratified random sampling was carried out on our dataset with training set having 65\% of data(119,155 images), 5\% in the validation set(11,147 images) and 30\% in the test set(55,848 images).
The Resnet34 classification model and the Seq to Seq model trained on our Atlas dataset
achieved an overall micro f-score of 92\% and 90\% respectively. A comparison of the f-scores of both the benchmark models over support size of leaf categories is shown in Figure
\ref{fig:metrics_comparison}. Though we observe that the classification model's performance is better than Seq to Seq model, we believe the reason is that we have only 52 categories at the moment.
As the number of categories increases, the structure in the taxonomy can be leveraged better using Seq to Seq model. In addition to Seq to Seq models predicting the category paths, it also explains the reason behind the predictions which is shown in Figure \ref{fig:sample_predictions}. 

\cite{li2018don} claim that using Seq to Seq model for product categorization helps to identify new category paths in the taxonomy. However, in our experiments, we have observed that all the new category paths that are generated by the Seq to Seq model are not always valid. In our case our model generated 5 new category paths which are shown in Table \ref{tab:new_categories} out of which we found only 2 to be valid. Therefore, a manual inspection of newly created category paths is needed to filter out the category paths which could be used to enrich the taxonomy.

\begin{table}
\small
\begin{tabular}{|l|l|}
\hline
\caption{Valid and invalid category paths created by Seq to Seq model}
  \label{tab:new_categories}
\textbf{Valid Category paths}                                                    & \textbf{Invalid Category paths}  \\ \hline
Women\textgreater Western Wear\textgreater Blazers\&Suits & Men\textgreater Western Wear\textgreater Dresses   \\ Women\textgreater{}Western Wear\textgreater Jackets        & Men\textgreater Western Wear\textgreater Tanktops \& Camisoles   \\
& Women\textgreater Inner Wear\textgreater Shorts \\ \hline
\end{tabular}
\end{table}

\section{Conclusion and Future Works}
\label{sec:conclusion}
This paper introduces \textbf{\textit{Atlas}}, a fashion apparel dataset with 186,150 apparel images along with their corresponding product titles. We have open-sourced the code base and the procedure to build the dataset of images and their taxonomy. We have proposed two benchmark models using classification and Attention based Sequence approaches to predict product taxonomy. 

In the future, we plan to extend our \textbf{\textit{Atlas}} dataset by adding more categories and products thereby increasing the total number of category paths. We would avoid the generation of invalid category paths in our Seq to Seq model by considering the taxonomy structure while decoding and explore Transformer Networks instead of Recurrent Neural Networks (RNN).

\section{Acknowledgements} The First Author Venkatesh Umaashankar worked extensively in the problem of Product Categorization using text attributes during his tenure at Indix. He thanks Krishna Sangeeth, Sriram Ramachandrasekaran, Anirudh Venkataraman, Manoj Mahalingam, Rajesh Muppalla and Sridhar Venkatesh for their help and support. 

\bibliographystyle{splncs04}
\bibliography{paper}
\end{document}